\documentclass{jimis-en}

\pdfoutput=1 

\usepackage{array}
\usepackage{pgfplots}
\usepackage{graphicx}
\usepackage[utf8]{inputenc}
\usepackage{esvect}
\usepackage{tabularx}
\usepackage{longtable}

\newcommand{\eg}{\textit{e.g.}}

\newcommand{\mic}{\mu\textnormal{m}}

\newcommand{\mmic}{\;\mic}
\newcommand{\ngc}[1]{NGC$\,$#1}

\title{Searching for Carriers of the Diffuse Interstellar Bands Across 
       Disciplines, using Natural Language Processing}
       
\author[1,2]{Corentin VAN DEN BROEK D'OBRENAN}
\author[1]{Frédéric GALLIANO}
\author[2]{Jeremy MINTON}
\author[2]{\\ Viktor BOTEV}
\author[*2]{Ronin WU}
\affil[1]{Université Paris-Saclay, Université Paris Cité, CEA, CNRS, AIM, 91191, Gif-sur-Yvette, France}
\affil[2]{Iris AI, Bekkestua, Norway} 
\corrauthor{ronin@iris.ai}

\doi{10.18713/JIMIS-ddmmyy-v-a}{}
\review{Jour Mois-en-lettres Année}{Jour Mois-en-lettres Année}
\publication{N}{AAAA}

\issue{thème interdisciplinaire}
\editors{Prénom-$1$ Nom-$1$, Prénom-$2$ Nom-$2$, Prénom-$k$ Nom-$k$...}

\begin{document}
\maketitle

\abstract{
The explosion of scientific publications overloads researchers with information.
This is even more dramatic for interdisciplinary studies, where several fields need to be explored.
A tool to help researchers overcome this is \textit{Natural Language Processing} (NLP): a machine-learning~(ML) technique that allows scientists to automatically synthesize information from many articles.
As a practical example, we have used NLP to conduct an interdisciplinary search for compounds that could be carriers for \textit{Diffuse Interstellar Bands} (DIBs), a long-standing open question in astrophysics.
We have trained a NLP model on a corpus of 1.5 million cross-domain articles in open access, and fine-tuned this model with a corpus of astrophysical publications about DIBs.
Our analysis points us toward several molecules, studied primarily in biology, having transitions at the wavelengths of several DIBs and composed of abundant interstellar atoms.
Several of these molecules contain chromophores, small molecular groups responsible for the molecule's colour, that could be promising candidate carriers.
Identifying viable carriers demonstrates the value of using NLP to tackle open scientific questions, in an interdisciplinary manner.}


\section{Introduction}

A prerequisite to interdisciplinarity is the ability of researchers in a given field to explore the literature of other fields and easily extract relevant information.
In particular, finding similar concepts in two different fields, adapting methods from one field to another, or re-purposing data acquired in an unrelated field are all potentially fruitful approaches for scientists to make discoveries. 
It is not unlikely that clues to various open questions in the global scientific literature currently exist, but looking for them is like searching for a needle in the proverbial haystack.
However, as the scientific knowledge expands exponentially~\citep{densen11}, human ability to keep up with both sorting and navigating diminishes quickly.
It is humanly impossible to read all published articles, and keyword searches that do not include contextual semantic meaning or conceptual reasoning are extremely limited. 
Fortunately, tremendous progress has recently been made in the automatic analysis of written documents. 
\textit{Natural Language Processing} (NLP) is a branch of linguistics that leverages the statistical properties of a given corpus with machine learning (ML) methods to explore the semantic relationships between texts~\citep{manning99}.
It is being used to extract information, to draw parallels between problems and to formulate new research directions. 
It is aimed at solving information overload. 
Recently, ML techniques have been used to generate scientific hypotheses in many scientific domains, such as the drug repositioning / discovery research~\citep{hastings2012,lamurias19}.
Going beyond the boundary of disciplines, we pioneer the use of ML techniques for hypothesis generation from cross-domain literature.
It is a practical method, quickly becoming popular, that will have an important role in interdisciplinary studies in the coming years.

The present article discusses the results of a collaboration between astrophysicists and computational linguists.
NLP techniques have already been applied to astrophysics to outline research priorities \citep{thomas22} and to thoroughly search the literature \citep{kezendorf19,grezes21}.
In these recent studies, NLP is used as an exploratory tool that can be used to help refine a project.
Here, we take a step further and use NLP as a primary research tool, with the hope of making actual discoveries.
We have applied it to the long-lasting question of the origin of the \textit{Diffuse Interstellar Bands} (DIBs). 
DIBs are ubiquitous spectral absorption features observed at visible wavelengths, along the sightline to stars in the Milky Way \citep{hobbs09}. 
They were discovered a century ago, but the chemical composition of their carriers is still unknown. 
Thus, we have trained a NLP model on a cross-domain corpus, which includes astrophysical publications about DIBs, then explored other fields, such as biochemistry, where relevant molecules could have been studied.
We start this article by presenting the astrophysical context of our study, in Section~\ref{sec:DIBs}.
In Section~\ref{sec:NLP}, we review the NLP techniques used in this study.
We then discuss how we have applied NLP to address the origin of DIBs, in Section~\ref{sec:project}.
The astrophysical relevance of the molecules we have found is assessed in Section~\ref{sec:results} and our results are summarized in Section~\ref{sec:summary}.

\section{Diffuse Interstellar Bands}
\label{sec:DIBs}

\begin{figure}[htbp]
\centering
\includegraphics[width=\textwidth]{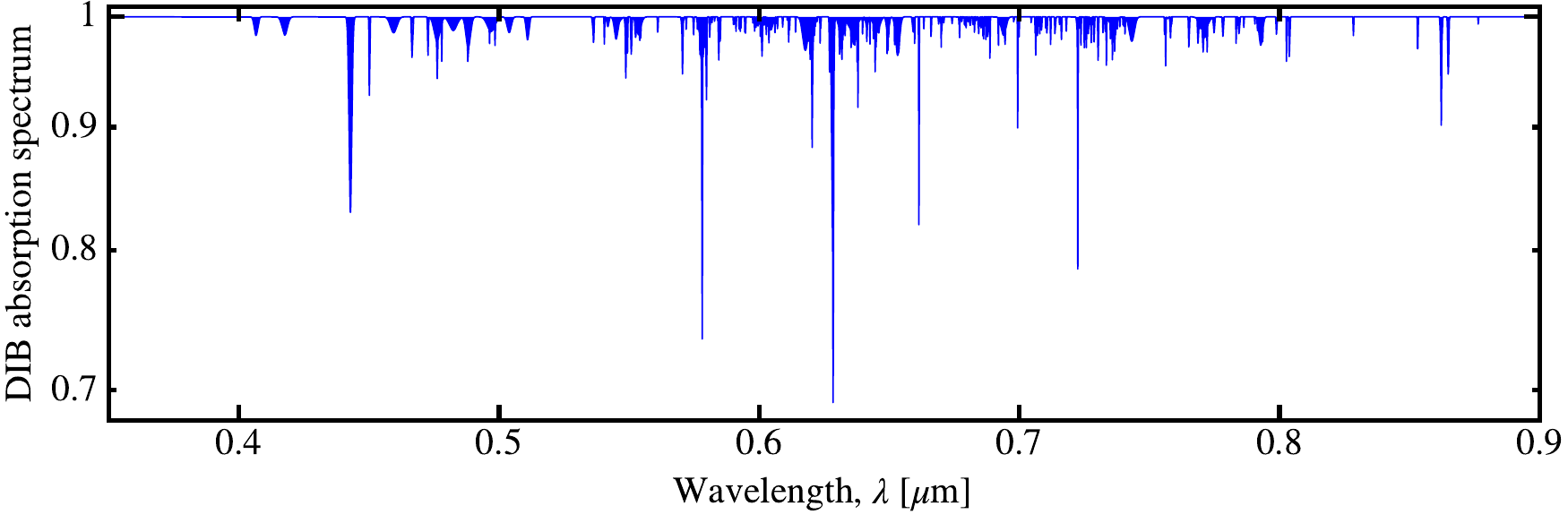}
\caption{Synthetic absorption spectrum of the most prominent DIBs.
         The parameters of each feature (position, width and depth) come from the observational compilation by \citet{jenniskens94}.}
\label{fig:dibs}
\end{figure}

The object of our study, the \textit{Diffuse Interstellar Bands} (DIBs), are spectroscopic absorption features that are ubiquitously observed in the Milky Way, our own galaxy \citep[][for reviews]{hobbs09,jones16b}.
They appear as a forest of bands along the line of sight towards stars, in the visible electromagnetic spectrum (wavelengths, $\lambda\simeq0.4-0.8\mmic$), sometimes extending up to the near-infrared domain (up to $\lambda=2\mmic$).
Figure~\ref{fig:dibs} shows a synthetic DIB absorption spectrum.
They originate from the \textit{InterStellar Medium} (ISM).

\subsection{The Interstellar Medium}

\begin{figure}[htbp]
\centering
\includegraphics[width=\textwidth]{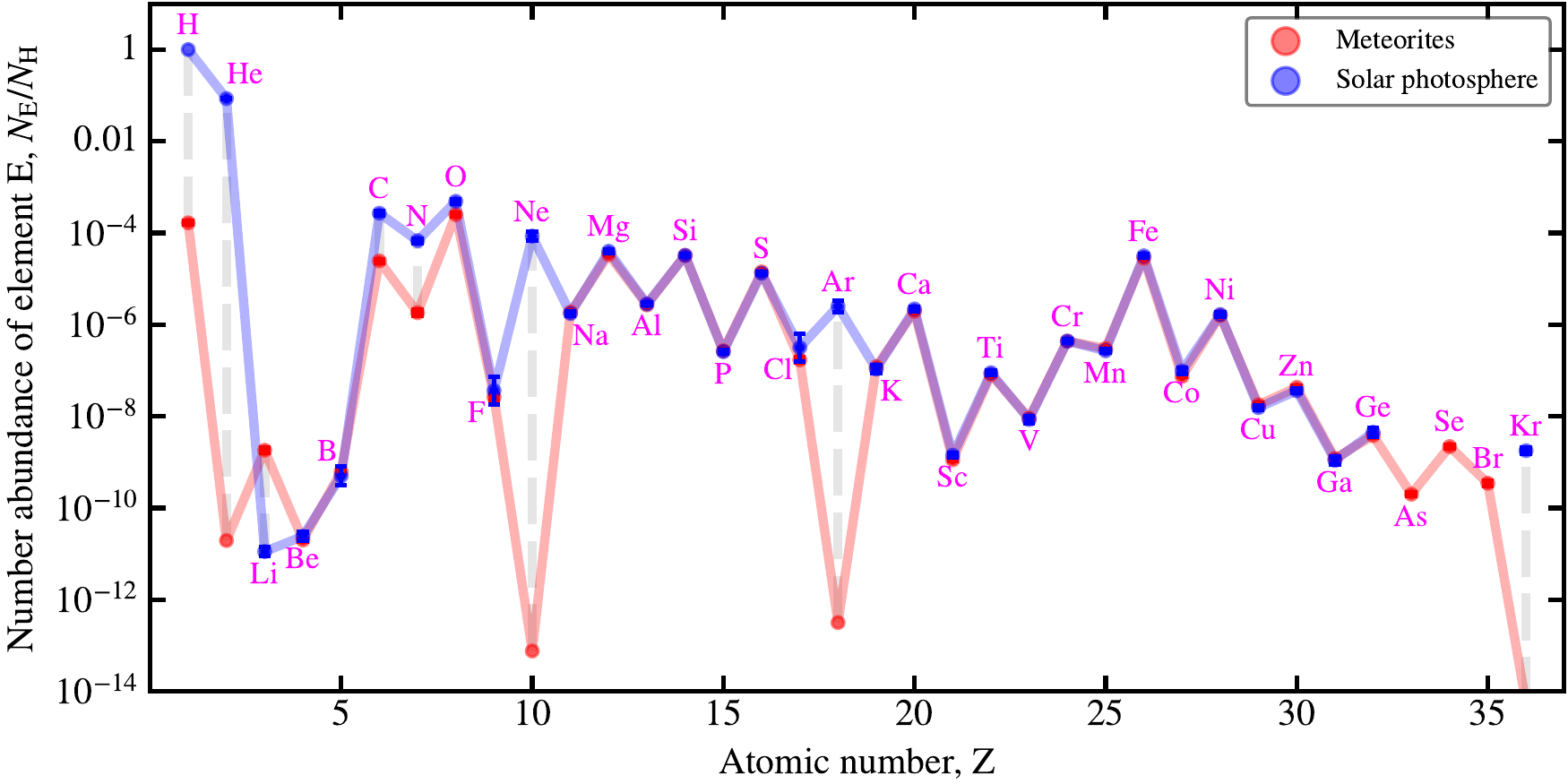}
\caption{Elemental abundances in the Solar system, from the data in 
         \citet{asplund09}. 
         Those are \textit{number} abundances (\textit{i.e.} number of atoms relative to hydrogen).
         The blue dots represent measures made in the Solar photosphere, through absorption spectroscopy, and the red dots correspond to the chemical composition of chondrite meteorites.
         Both are in good agreement, except for the lightest, most volatile elements, which do not remain trapped in meteorites.
         These abundances are representative of the ISM in our Solar neighborhood, and are even used as a reference when studying other galaxies.}
\label{fig:abundance}
\end{figure}

Since DIBs are interstellar features, we first need to review the general properties of the ISM.
Broadly speaking, the ISM is constituted of all the matter filling the volume of a galaxy between the stars.
This matter is essentially gaseous, but about half a percent of its mass is made of small solid particles, the dust grains \citep[][for textbooks]{tielens05,Draine}.
The elemental composition of the ISM is $74\,\%$ hydrogen, $25\,\%$ helium, and the remaining $1\,\%$ contains all the heavier elements \citep{asplund09}.
Figure~\ref{fig:abundance} shows the abundances, in the Solar system, of the first elements in the periodic table.
We see that besides hydrogen and helium, the two most abundant species are carbon and oxygen.
This puts some constraint on the most abundant molecules that can be formed, and this will be important for our serendipitous search.

  \subsubsection{The Phases of the ISM}

The ISM is a highly heterogeneous medium \citep[\eg\ Chap.~III.3 of][]{galliano22}.
Half of the volume of our galaxy is filled with a permeating hot ionized gas with a very low density (temperature, $T=10^6$ K; density, $n\simeq3\times10^{-3}\;\textnormal{cm}^{-3}$).
This phase is heated by the shock waves from supernova explosions.
The rest of the volume covers a wide range of density, temperature and atomic state. 
The coldest and densest interstellar regions are called molecular clouds.
As their name indicates, the elements in these clouds have combined to form molecules, majoritarily H$_2$ and CO.
Their high density (up to $n\simeq10^6\;\textnormal{cm}^{-3}$) allow them to be shielded from the stellar radiation and to reach low temperatures ($T\simeq10$~K).
Between these two extremes, there are nine orders of magnitude in density and five in temperature, unlike anything we can find on Earth.
DIBs are found preferentially in the diffuse ISM, and disappear in dense regions \citep[\eg][]{lan15}, although some specific DIBs are found in diffuse molecular clouds \citep[density $n\simeq10^2\;\textnormal{cm}^{-3}$;][]{thorburn03}.

  \subsubsection{Interstellar Molecules}
  \label{sec:molecules}

As we will see in Section~\ref{sec:DIBcarriers}, DIB carriers are likely large molecules.
Until now, more than 200 individual molecules have been identified in space \citep{mcguire18}.
The first one to be discovered was CH, in the 1930s \citep[][for an historical review]{herzberg88}.
Although this first observation was performed in the visible domain, the majority of the subsequent detections were achieved at radio wavelengths, through the various rotational lines\footnote{Rotational lines arise from transitions between different values of the quantum angular momentum of the molecule.} of the molecules.
Several new molecules containing more than six atoms are now discovered each year.
These large molecules all contain carbon atoms.
They are thus called \textit{Complex Organic Molecules} \citep[COMs;][for a review]{herbst09}.
Even branched molecules, which were believed to be too brittle to survive the harsh interstellar environment, have been detected \citep{belloche14}.

\begin{figure}[htbp]
  \centering\
  \begin{tabular}{cc}
    \includegraphics[width=0.63\textwidth]{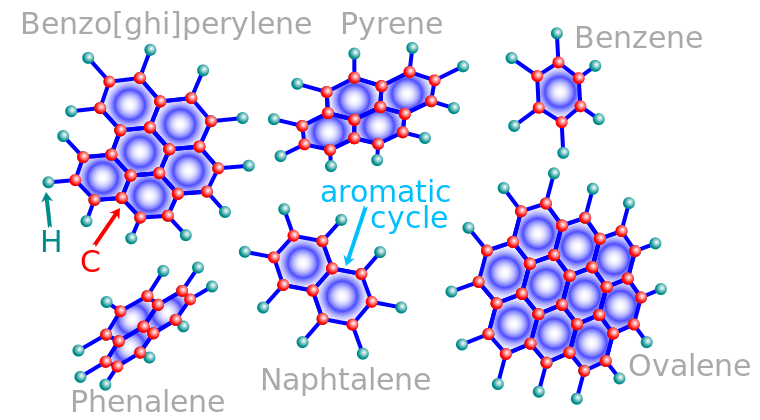} &
    \includegraphics[width=0.33\textwidth]{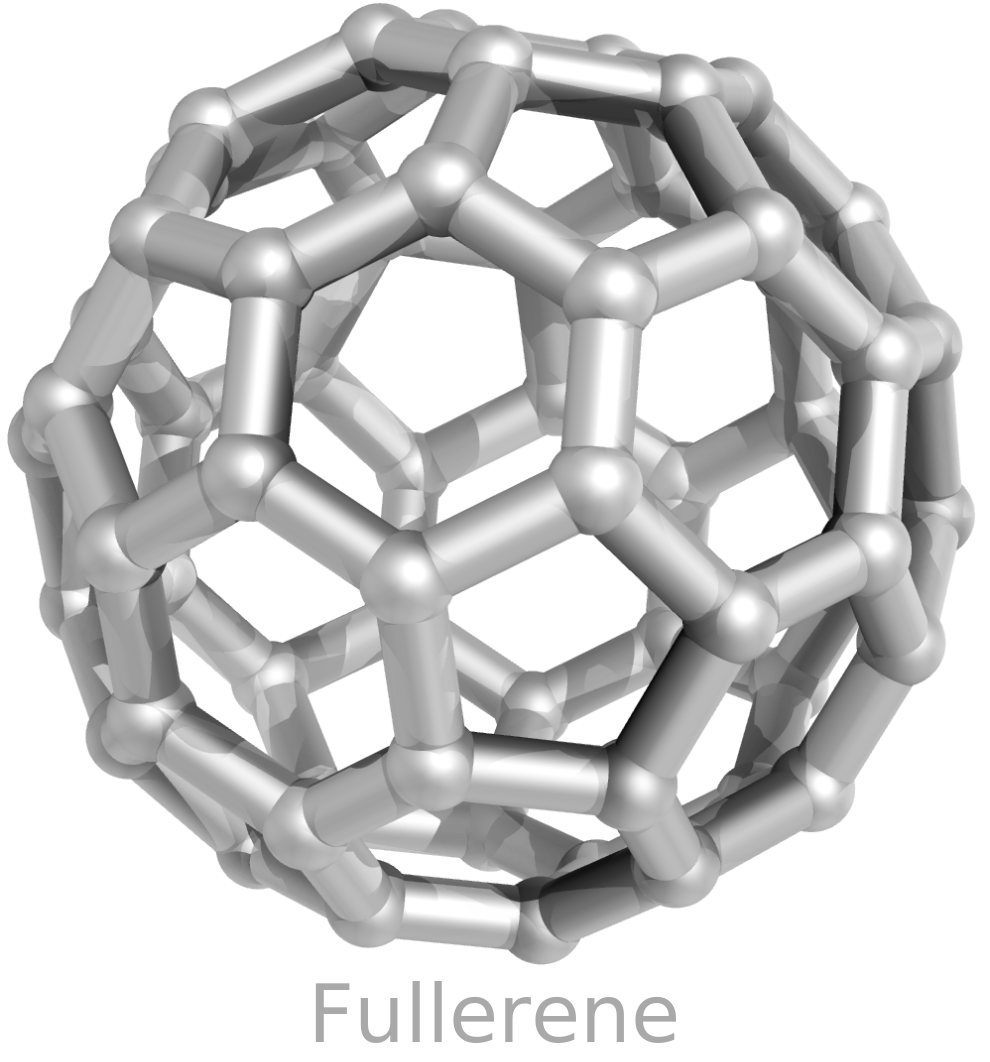} \\
  \end{tabular}
  \caption{The PAH molecular family.
           On the left side, we show six different PAHs.
           Hydrogen atoms are represented in cyan, and carbon atoms, in red.
           On the right side, we show the buckminsterfullerene, which is a spherical molecule, composed of 60 carbon atoms.
           Credit: the left figure is adapted from \citet{galliano22}; the fullerene image is from \href{https://commons.wikimedia.org/wiki/File:Fullerene.png}{Yassine Mrabet}, licensed under \href{https://creativecommons.org/licenses/by/3.0/deed.en}{CC BY 3.0}.}
  \label{fig:PAHs}
\end{figure}

\begin{figure}[htbp]
  \includegraphics[width=\textwidth]{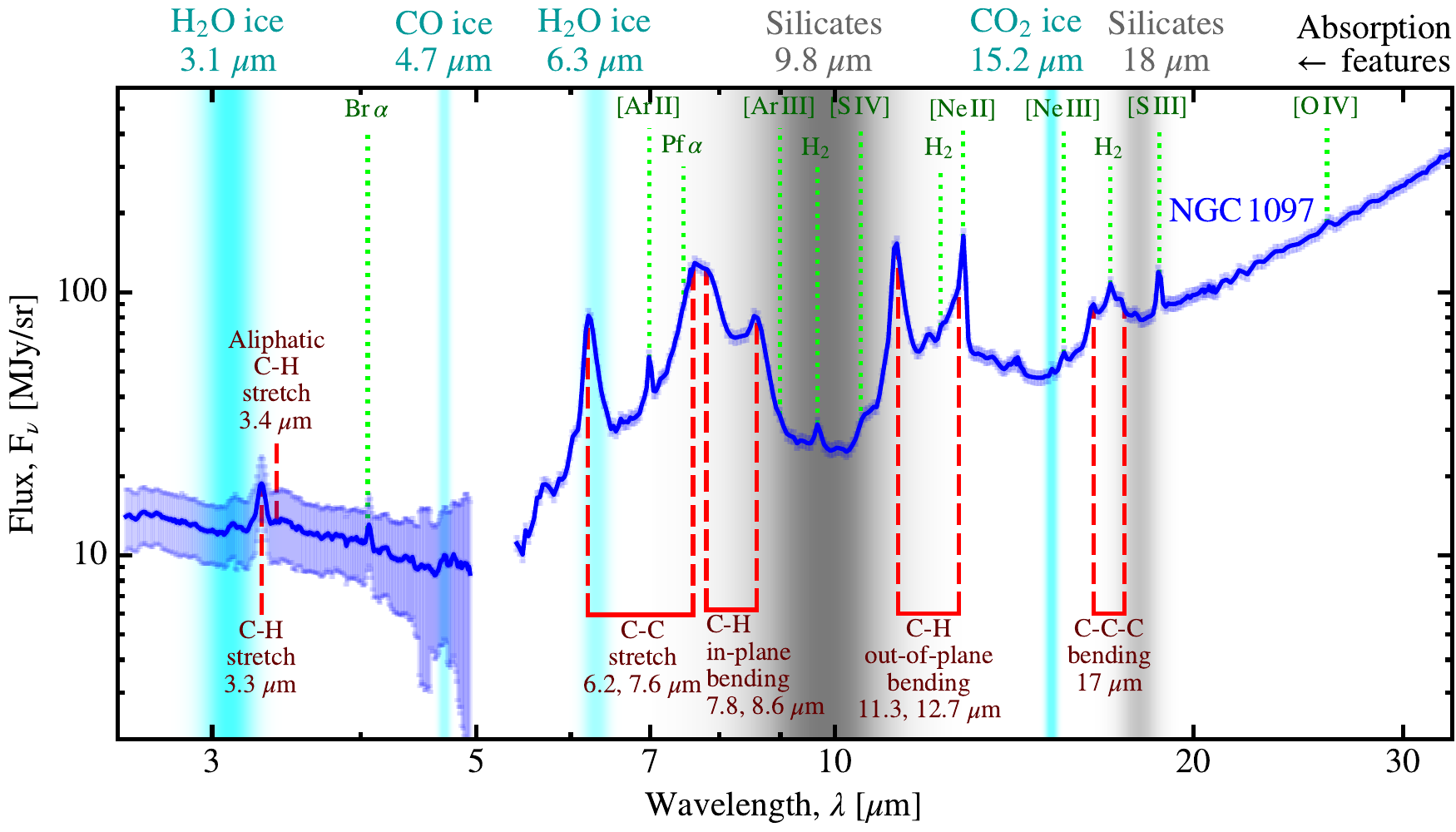}
  \caption{Mid-infrared spectrum of the galaxy \ngc{1097}.
           The blue line, with error bars, is the observed spectrum of the central region of this object.
           We have highlighted the main spectral features:
           on top, we have indicated the main silicate and ice interstellar absorption bands;
           in green, we have shown the position of the brightest gas lines;
           in red, we have pointed towards the brightest PAH features and have noted which vibrational C--C or C--H mode they correspond to.
           Credit: figure adapted from \citet{galliano18}.}
  \label{fig:specMIR}             
\end{figure}

An important class of organic molecules is \textit{Polycyclic Aromatic Hydrocarbons} (PAHs).
They are constituted of several \textit{aromatic cycles}, which are hexagonal structures made of carbon atoms, such as in benzene, with peripheral hydrogen atoms (Figure~\ref{fig:PAHs}).
This species was introduced, in astrophysics, to provide an interpretation for the bright mid-infrared features (in the spectral range $\lambda=3-20\mmic$) observed ubiquitously in the ISM and in galaxies (Figure~\ref{fig:specMIR}).
The vibrational modes of the C--C and C--H bonds in PAHs, which have similar resonance frequencies across the PAH family, indeed provide a good account for these mid-infrared bands \citep[][for a review]{tielens08}.
This however forbids the identification of individual PAHs, as these broad mid-infrared bands arise from the mixture of several molecules.
Only a few individual PAHs have unambiguously been detected, either because they have very peculiar features due to an atypical structure, such as fullerenes \citep[Figure~\ref{fig:PAHs};][]{cami10}, or, recently, through their rotational lines \citep{mcguire21}.

  \subsubsection{Interstellar Dust Grains}

When the number of atoms in a molecule becomes large, its resonance features become broader and a noticeable continuum arises \citep[\eg\ Chap.~1 of][]{galliano22}.
This is because, with a large number of atoms, we are entering the solid-state realm.
Large molecules are at the interface with dust grains, and there is probably a continuity between both in the ISM, although there is no well-defined limit between these two categories.

Interstellar dust grains are small solid particles with radii ranging from $\simeq3$~\AA\ to 300~nm \citep[\eg][for a review]{draine03}.
They account for about half of the mass of heavy elements in the ISM, thus about half a percent of its total mass.
Yet, these grains, which are predominantly silicate and carbonaceous compounds, have a very important role.
In a quiescent galaxy, such as the Milky Way, they absorb about 25$\,\%$ of the stellar power, in the ultraviolet (UV) and visible range, and re-emit it thermally in the infrared \citep[\eg][]{bianchi18}.
In regions of massive star formation, this fraction can go up to 99$\,\%$.
These regions are thus totally opaque to visible photons and can be studied only through their infrared radiation, emitted by the dust.
Grains are also the catalysts of several chemical reactions, including the formation of H$_2$, the most abundant molecule in the Universe \citep[\eg][]{bron14b}.

When looking at a star in the Milky Way, the dust present along the line of sight extincts its radiation. 
It means that a fraction of the stellar light is either absorbed by grains or scattered in another direction.
Dust grains are so well-mixed with the gas in the ISM that the extinction amplitude (usually quoted in the visual band, V, at $\lambda=0.55\mmic$) is considered as a reliable tracer of interstellar matter.

  \subsubsection{The Lifecycle of a Galaxy}

Finally, it is important to understand that the ISM is a dynamical environment, constantly evolving.
It is indeed the fuel of star formation.
Stars form by the gravitational collapse of molecular clouds.
Once ignited, the most massive stars, which are also the brightest, blow their interstellar cocoon away and ionize their surroundings.
At the end of their lifetime, stars eject in the ISM fresh heavy elements that they have formed in their core by nucleosynthesis.
The more a galaxy is evolved, the more tenuous and rich in heavy elements its ISM is.

\subsection{The Uncertain Nature of DIBs}
\label{sec:DIBcarriers}

The first DIBs were discovered exactly one century ago by \citet{heger22}, but the physical nature of their carriers largely remains a mystery, today.
Their interstellar origin was demonstrated by \citet{merrill34}.
Their intensity is indeed correlated with the dust extinction amplitude and independent of the intrinsic properties of the background stars.
Over 500 DIBs have been detected so far, in the ISM \citep{fan19}.
DIBs are also detected in external galaxies, such as the Magellanic clouds \citep[][for a review]{galliano18}.

  \subsubsection{Constraints on the Nature of the DIB Carriers}
  \label{sec:constraints}

As absorption features, DIBs must originate from the transition of an atom or a molecule, between two of its quantum energy levels.
The intrinsic line width of DIBs, which is typically $\simeq1$~\AA, excludes that they originate from free-flying atoms.
They must come from molecules with a few to $\simeq100$ atoms \citep{macisaac22}.
Their strength and their ubiquity also tells us that they have to be made with the most abundant atoms in the ISM, mainly H, O, C, N.
A last constraint is that, due to their presence in the diffuse ISM, a medium permeated with UV photons, these molecules need to be rather compact, to be more resilient.
The branched molecules we have mentioned in Sect.~\ref{sec:molecules} are only found in dense molecular clouds, well protected from UV radiation.

  \subsubsection{Identification of Buckminsterfullerene}

The only molecule to date, unambiguously identified as a DIB carrier is the \textit{buckminsterfulerene}\footnote{Fullerenes constitute a family of compact closed-mesh carbon compounds. Buckminsterfullerene is the variety with formula C$_{60}$.} \citep{campbell15,walker15}.
The cation of this molecule, C$_{60}^+$, can account for two, possibly four DIBs, at near-infrared wavelengths.
Notice that this molecule satisfies all the constraints we have listed in Sect.~\ref{sec:constraints}.
The fact that this molecule had been detected beforehand, in a planetary nebula, \textit{via} its mid-IR features \citep{cami10}, makes this identification trustworthy.
It is possible that other COMs, and probably other PAHs, are DIB carriers.
There are however variations of the relative strengths of DIBs, across sightlines \citep{herbig95}.
This indicates that DIB carriers likely come from a diversity of molecules whose relative abundance varies with the environment.

  \subsubsection{The Significance of DIB Identification}
  
 DIBs are the last large class of interstellar spectral features that are still unidentified.
 The fact that, a hundred years after their discovery, less than one percent of these features has been identified, shows that this is an arduous challenge.
 This is however an important question, as spectroscopy is historically what transformed astronomy into astrophysics \citep{hearnshaw14}.
 This is because spectroscopy allows us to identify atoms and molecules from a distance, measure their charge state, temperature, density and abundance, that we could learn so much about the Universe.
 Identifying the carriers of the DIBs would therefore be a major breakthrough in our understanding of the ISM.
 It could help us unlock the complexity of interstellar chemistry and provide a wealth of diagnostics of the physical conditions where these bands are observed.
 
 Astrophysicists have been stuck by this problem, because its answer lies in the wide diversity of potential molecules.
 Only a handful of these molecules can be measured in the laboratory or computed theoretically.
 Teams working on these identifications have limited resources.
 Yet, since we have seen that DIB carriers are likely large organic molecules, it is possible that some of these molecules have been studied in other fields, such as biochemistry or biology.
 We have thus performed a serendipitous, interdisciplinary search for these molecules, using the technique of natural language processing.

\section{Natural Language Processing}
\label{sec:NLP}

NLP is a sub-discipline of machine-learning, that combines linguistics and artificial intelligence to allow computers to interface with human language.
In this application specifically, it is used to extract data from natural language documents: academic papers. 

\subsection{Machine learning and neural networks}

\textit{Machine-Learning} (ML) is a methodology for producing computer programs whose main decision making is determined by data, not directly coded by a computer scientist.
This methodology enables us to perform automated decision making too detailed for a human to design by using a set of examples too numerous for a human to fully perceive.
Done correctly, the decision making should remain valid for data not explicitly included in the set of training examples~\citep[][for a text book]{goodfellow16}.
\textit{Neural Networks} (NNs) are a popular algorithm for many modern ML applications.
Modelled on an organic brain, NNs are comprised of artificial neurons that perform a non-linear operation on a signal and propagate the result as a signal to the next neurons. When that signal has passed through the entire network of neurons, it can be interpreted as a meaningful prediction.
Each neuron's behaviour is parameterized.
Corrections are made to iteratively update the neurons' parameters during a dedicated training process by propagating a prediction error backwards through the network and calculating the error gradient with respect to each parameter.
Neurons are typically arranged in layers because the corresponding vectors of parameters allow efficient matrix arithmetic to be used to perform the necessary operations.

\subsection{Word-embeddings}

Word-embeddings are numerical representations of words, useful for computation, and NNs are a common tool for producing and using word-embeddings.

In recent years, development of word-embeddings models has taken big steps forward from NNs with few layers, such as the Word2Vec~\citep{mikolov13b}, Glove~\citep{pennington14}, and FastText~\citep{bojanowski17} models, to complex models that require enormous volumes of data to train, such as ELMo~\citep{elmo} and BERT~\citep{devlin2019}.

\subsubsection{Concepts}
Word-embeddings are underpinned by the distributional hypothesis: that words with similar co-occurrence distributions have similar meanings or "a word is characterized by the company it keeps"~\citep{firth1957}.
Techniques leveraging this hypothesis are able to represent semantic properties of words and capture meaningful syntactic and semantic regularities.
Often, regularities are observed as constant vector offsets between pairs of words sharing a particular relationship.
For example, it has been demonstrated that the Word2Vec model in \citet{mikolov13a}  can learn the male/female relationship, which is expressed as the vector expression, $\vv{king} - \vv{man} + \vv{woman}$ producing a vector very close to $\vv{queen}$.
Such behaviour makes word-embedding models an effective tool for a number of NLP applications such as identifying contextual synonyms, ranking keywords, and computing similarities between millions of documents~\citep{botev17}.

\subsubsection{The Word2Vec Model}
\label{ref:w2v}

Given the large volume of publications to be processed in this study, we opt to use the \texttt{gensim} library~\citep{rehurek10} implementation of the Word2Vec Continuous Bag-of-Words (Word2Vec-CBOW) model  for it's computational speed.
The Word2Vec model is a simple architecture of two sets of embeddings where one encodes input words that should approximate output words encoded by the other. This will draw similar words together and push dissimilar words apart within the embedding space. The two variants, CBOW and skip-gram, swap the target word and context words as input and output.
The CBOW model uses context words that appear in a fixed window around the target word as input to predict the target word. See \citet{mikolov13b} for details about context word sampling, negative-word sampling and the loss-function used.

In order to obtain a model trained on the cross-domain scientific literature, we compile a generic corpus of 1.5 million open access English articles across all domains from the CORE service~\citep{knoth12}.
To ensure the articles cover a diverse range of domains, we randomly select these articles out of a pool of 20 million English articles hosted by the CORE service with any of 24 keywords.
The keywords used in our selection and the resultant numbers of articles are listed in Table~\ref{tab:genre_keywords}.
\begin{center}
\begin{longtable}{|l|l|}
\caption{Genre keywords used to select articles in July 2021} \label{tab:genre_keywords} \\
\hline \multicolumn{1}{|c|}{\textbf{Keyword}} & \multicolumn{1}{c|}{\textbf{\#\ of articles}} \\ \hline 
\endhead
\hline
\endfoot
"starburst" $\land$ "galaxy" & 55k  \\
"quantum" $\land$ "field" $\land$ "theory" & 50k \\
"artificial" $\land$ "intelligence" & 80k \\
"nanotechnology" & 75k  \\
"surface" $\land$ "science" & 25k \\
"microbiome" & 75k   \\
"computer" $\land$ "architecture" & 75k \\
"game" $\land$ "theory" & 25k  \\
"bioinformatics" & 25k \\
"quantum" $\land$ "computation" & 75k \\
"cognitive" $\land$ "behavioral" $\land$ "therapy" & 40k \\
"obsessive" $\land$ "compulsive" $\land$ "disorder" & 40k  \\
"renaissance" & 20k  \\
"crystallography"  & 75k \\
"dielectric" & 75k  \\
"Ising" & 25k  \\
"Markov" $\land$ "chain" & 75k  \\
"thin" $\land$ "film" & 75k  \\
"cancer" & 80k \\
"corrosion" & 75k  \\
"neural" $\land$ "network" & 75k  \\
"geochemistry" & 15k \\
"HIV" & 75k  \\
"renewable" $\land$ "energy" & 75k  \\
\end{longtable}
\end{center}
Due to the resource limitation in this pilot study, we focus on the abstracts, which are usually concise but still contain essential information of the article.
We then train a generic Word2Vec-CBOW model of 100 dimensions on this corpus for ten epochs with a window size of five: where the dimensionality is the size of the vector embedding for each word, one for input and one for output, and the epochs are the number of times the entire corpus is repeated during training, and the window size specifies the region around the target word from which context words can be selected.
While hyperparameter optimization was not performed on this specific dataset, the choice of dimension was based on previous experiments \citep{yin2018} and the \texttt{gensim} library's recommended value for CBOW models.
We observe that training over the corpus for ten epochs showed thorough convergence of validation loss.
In the following section, we explain how this model is used in our analysis.

\subsection{Extracting physical quantities from scientific documents}
\label{sec:unit}
As the main objective of this study is to discover possible carriers of DIBs in the literature outside of the astrophysics domain, recognition of physical quantities is an essential step in the corpus processing.
A physical quantity in the corpus is defined as a numeric value that appears with a physical unit, such as ``$1500~\mathrm{\AA}$''.
However, recognizing physical quantities from existing documents is not a trivial task. 
For example, MWC 349A, is a member of the double star system, MWC~349~\citep{gvaramadze12}, and should not be recognized as a physical quantity of $349\;\mathrm{\AA}$.
Furthermore, there are numerous ways to denote one physical quantity.
For example,  $1500~\mathrm{\AA}$ and $0.15~\mathrm{\mu m}$ are equivalent and should be recognized as such by the model.

To identify all relevant wavelengths mentioned in the generic corpus, we implement a quantity recognition algorithm which includes three major components:
\begin{enumerate}
    \item A regular expression~(RegEx) algorithm that identifies all numerical values, such as $28.4\times10^3$, $28,400$ and $(28.4\pm0.1)\times10^3$.
    \item A unit-parsing algorithm that is based on the open source library Pint (version 0.19.1)\footnote{\href{https://pint.readthedocs.io/en/stable/}{https://pint.readthedocs.io/en/stable/}}.
    \item A unit-disambiguation algorithm that assigns probabilities to ambiguous units, such as angstrom (\AA) and ampere (A), by comparing the unit predicted by the generic Word2Vec model.
\end{enumerate}
We detail the unit-disambiguation algorithm in the following.

With the RegEx and unit-parsing algorithms listed above, we mask the identified physical quantities by numbers and units in the articles of the generic corpus.
Physical quantities with ambiguous units are masked by all candidate units, for example, $1500~\mathrm{\AA}$ is masked as ``NUM--Angstrom--Ampere''.

Using this model, we compute a context vector, which is the average vector from a window of 5 words around the physical quantity excluding itself, for each physical quantity with an ambiguous unit.
The choice of window size, $5$, is to be consistent with the hyperparameters used in the Word2Vec-CBOW model training discussed previously.
For each candidate unit assigned to the ambiguous unit, we then calculate the cosine similarity, which quantifies how parallel are two vectors, or in our case, how contextually-correlated are two words, between the calculated context vector to other units of the same dimension as the candidate unit.
The highest cosine similarity is then assigned to the candidate unit as its score to represent the ambiguous unit. 

\section{TACKLING DIBs WITH NLP}
\label{sec:project}

We now discuss how we apply this NLP model to the question of DIBs.
We start with the curation of the astrophysical literature corpus.

\subsection{Compiling the DIB corpus}

\begin{figure}[htb]
\centering
\includegraphics[width=\textwidth]{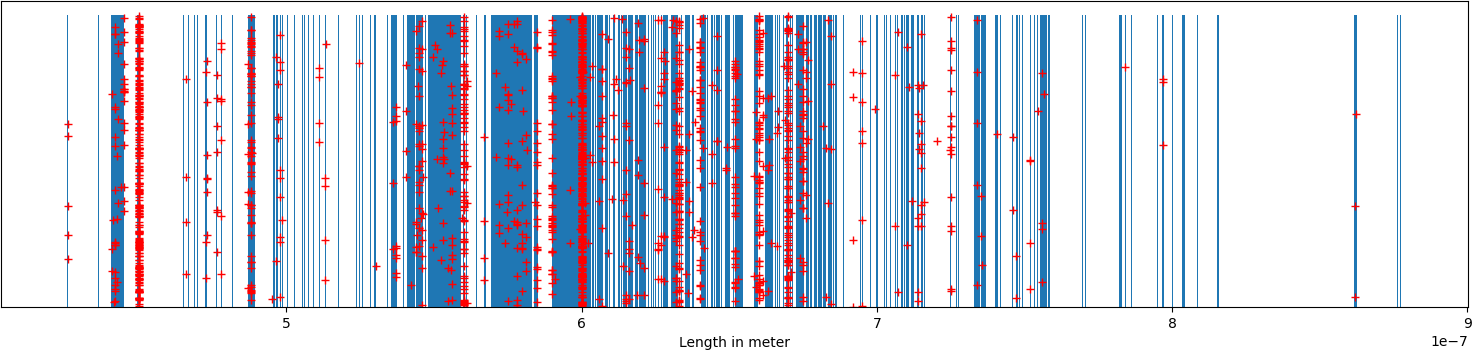}
\caption{Indication of identified known DIBs (blue bands) and identified physical quantities from the generic corpus that overlap with DIBs (red crosses). The y-axis is in arbitrary unit to guide the eyes on how identified physical quantities distribute through the DIBs bands.}
\label{fig:BruteForceSpectrum}
\end{figure}
To improve accuracy of the semantic relationships between words directly associated with DIBs and common words used in other domains, a specialized corpus for DIBs is required.
Our generic corpus of 1.5 million articles aims to cover cross-domain literature so will likely under-represent DIB articles making it inadequate for this purpose.
To overcome this under-representation, we compile a corpus specialized in DIBs to enhance the representations of DIB words in the word-embeddings model.

We first search for DIBs-related papers on NASA’s Astrophysics Data System\footnote{\href{https://ui.adsabs.harvard.edu/}{https://ui.adsabs.harvard.edu/}} and look for their relevant papers with Iris.ai's Explore Tool\footnote{\href{https://the.iris.ai/}{https://the.iris.ai/}}.
We then examine all articles appearing in the search and hand pick the articles that are directly DIBs-related.
The final selected DIB corpus then includes 939 articles.
We then search through these 939 articles to annotate all valid physical quantities in the texts that are directly associated with the DIBs catalog~\citep{jenniskens94}\footnote{\href{https://leonid.arc.nasa.gov/DIBcatalog.html}{https://leonid.arc.nasa.gov/DIBcatalog.html}}.
In total, there are 243 annotated wavelengths identified out of these articles.
The physical quantity recognition algorithm described in Section~\ref{sec:unit} correctly identified 203 and missed 40 of them.
The algorithm also wrongly identified 41 candidates that are not valid physical quantities.
The precision and recall values are summarized in Table~\ref{tab:precision_recall}.
These 243 annotated wavelengths, along with the full-width-half-maximum (FWHM), are marked as blue bands in Figure~\ref{fig:BruteForceSpectrum}.
\begin{table}[htb]
\caption{Precision and recall for the physical quantity recognition algorithm.}
\label{tab:precision_recall}
\centering
\begin{tabular}{|c|c|}
\hline
precision & recall \\
\hline
203/244 & 203/243\\
\hline
83.2\% & 83.5\%\\ 
\hline
\end{tabular}
\end{table}

\begin{figure}[htb]
\centering
\includegraphics[width=0.8\textwidth]{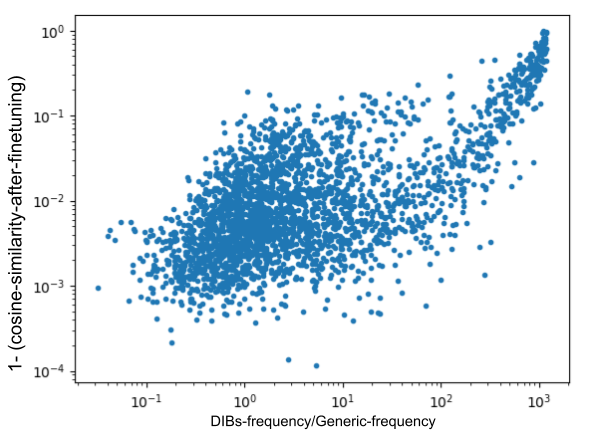}
\caption{This graph shows how tokens specific to the DIB corpus vary before and after the fine-tuning. One can see that tokens that are frequent in the DIB corpus yet infrequent in the genric corpus displace the most, while tokens that are infrequent in the DIB corpus yet frequent in the generic corpus do not displace much.}
\label{fig:cos_sim_updates}
\end{figure}

As many tokens that are specific to the DIB corpus are infrequent in the generic corpus, in order to well represent these tokens, it is important to fine-tune the generic Word2Vec-CBOW model.
The DIB corpus, with the disambiguated units, is then used to fine-tune the generic model.
During the fine-tuning process, the vectors of tokens that appear in the DIB corpus are updated and thus displaced with the new context words.
We observe that after training the generic model on the DIB corpus for ten epochs, the displacement of vectors stabilized.
As demonstrated in Figure~\ref{fig:cos_sim_updates}, in the fine-tuning process, vectors of tokens that are frequent in the DIB corpus and infrequent in the generic corpus are displaced the most while vectors of tokens that are infrequent in the DIB corpus yet frequent in the generic corpus are generally still.

\subsection{Using the unit quantity recognition}
We then apply the physical quantity recognition algorithm on the generic corpus.
Out of the 1.5 million articles, we identify $\sim2000$ physical quantities from $\sim20,000$ articles, that overlap with DIBs and mark their locations as red crosses in Figure~\ref{fig:BruteForceSpectrum}.
Some of these physical quantities describe molecular sizes, which are unrelated to DIBs, but others describe the wavelength of the energy transitions and are thus of direct interest to us.

In order to systematically filter the identified physical quantities down to the most relevant to DIBs, we use three filters to select candidate articles.
\begin{enumerate}
    \item In the DIBs wavelength range, $0.1 - 1\;\mu m$, physical quantities that are described in the unit of $\mu m$ in the generic corpus are often found to be associated to diameters or distances, such as the DNA lengths.
          On the contrary, physical quantities used to describe wavelengths in this range often appear in units of angstrom or nanometer.
          As a consequence, we discard quantities that are identified within this range that appear in $\mu m$.
    \item Known laser wavelengths are irrelevant for identifying DIB carriers so we discard any identified physical quantities that co-occur with the token "laser" or "light" within a window of 5 words.
    \item Using the DIBs-enhanced Word2Vec-CBOW model in the sentence window as the context where a physical quantity is identified, we compute a cosine similarity between the context vector and the physical-quantity vector.
          Articles of low~($<0.5$) cosine similarities are filtered out to remove false positives given by the unit-recognition algorithm.
\end{enumerate}
After applying filters, we manually screen through all 1932 candidate articles and identify twelve papers of high relevance, which are discussed in the following section.
\section{ASSESSMENT OF THE RESULTS}
\label{sec:results}

Our model points us toward twelve articles presenting spectroscopic measurements of molecules having transitions corresponding to some DIBs.
These findings are summarized in Table~\ref{tab:results}.
They now need to be scrutinized.

\subsection{The Results}

\setlongtables
\begin{longtable}{|l|l|l|l|}
\hline
\textbf{Article} & \textbf{Transitions} & \textbf{Closest DIB} & \textbf{Molecule}\\
\hline
\endhead
\citet{wakakuwa10} & $425$ nm & $425.90\pm0.01$ nm & 11-cis retinal\\
Butterfly eye pigment & $453$ nm & \textcolor{gray}{$450.18\pm0.02$ nm} & chromophore\\
 & $563$ nm & $563.50\pm0.01$ nm & within opsin \\
 & $620$ nm & $619.90\pm0.01$ nm & (H, C, O)\\
 & $640$ nm & $640.05\pm0.01$ nm & \\
\hline
\citet{dove95} & $560$ nm & $560.09\pm0.01$ nm & Chromophores\\
Coral pigment & $580$ nm & $580.66\pm0.01$ nm & within\\
 & $590$ nm & $590.06\pm0.01$ nm & pocilloporin\\
\hline
\citet{davies09} & $441.9\pm1.0$ nm & $442.82\pm0.17$ nm & Chromophores\\
Elephant shark eye pigment & $493.7\pm2.6$ nm & $494.74\pm0.01$ nm & within\\
 & $496.3\pm0.1$ nm & $496.39\pm0.01$ nm & proteins\\
 & $498.3\pm0.3$ nm & $498.21\pm0.01$ nm & (H, C, N, O)\\
 & $498.7\pm0.3$ nm & $498.74\pm0.01$ nm & \\
 & $504.1\pm1.0$ nm & $505.48\pm0.01$ nm & \\
 & $509.5\pm0.5$ nm & $509.21\pm0.01$ nm & \\
 & $510.1\pm0.2$ nm & $510.10\pm0.01$ nm & \\
 & $520.9\pm2.0$ nm & $521.79\pm0.01$ nm & \\
 & $534.2\pm1.0$ nm & $534.25\pm0.01$ nm & \\
 & $547.8\pm2.2$ nm & $548.08\pm0.01$ nm & \\
\hline
\citet{spady06} & $423$ nm & \textcolor{gray}{$425.90\pm0.01$ nm} & Retinal\\
Cichlid eye pigment & $456$ nm & \textcolor{gray}{$450.18\pm0.15$ nm} & chromophores\\
 & $472$ nm & $472.68\pm0.02$ nm & within\\
 & $518$ nm & $517.81\pm0.01$ nm & opsin\\
 & $528$ nm & \textcolor{gray}{$529.8\pm0.01$ nm} & \\
 & $561$ nm & $560.98\pm0.01$ nm & \\
\hline
\citet{wolfbeis00} & $488$ nm & $488.00\pm0.01$ nm & Valinomycin\\
Valinomycin &  &  & (H, C, N, O)\\
\hline
\citet{filosa01} & $695$ nm & $694.46\pm0.01$ nm & Amide II\\
Blood proteins &  &  & (H, C, N, O)\\
\hline
\citet{davies09b} & $501.0\pm0.1$ nm & \textcolor{gray}{$500.36\pm0.01$ nm} & Chromophores\\
Agnathan eye pigments & $535.5\pm3.3$ nm & $535.88\pm0.01$ nm & within\\
 & $544.1\pm5.0$ nm & $543.35\pm0.01$ nm & opsin\\
 & $554.3\pm2.0$ nm & $554.51\pm0.01$ nm & \\
 & $562.8\pm0.4$ nm & $563.50\pm0.01$ nm & \\
\hline
\citet{schoot-uiterkamp76} & $653$ nm & $653.65\pm0.01$ nm & Tyrosinase\\
Mushroom absorption & $755$ nm & $755.94\pm0.01$ nm & (H, C, O, Cu)\\
\hline
\citet{davies07} & $439$ nm & \textcolor{gray}{$436.39\pm1.10$ nm} & Chromophores\\
Lamprey eye pigment & $492$ nm & \textcolor{gray}{$494.74\pm0.01$ nm} & \\
 & $497$ nm & $496.91\pm0.01$ nm & \\
\hline
\citet{marechal07} & $445$ nm & $442.82\pm1.69$ nm & Heme intermediate\\
Nitric-oxide synthase &  &  & (H, C, N, O, Fe)\\
\hline
\citet{remigy03} & $420$ nm & \textcolor{gray}{$425.90\pm0.01$ nm} & Heme-type\\
Bacterium cytochrome & $525.2$ nm & $525.18\pm0.01$ nm & molecule\\
 & $545.4$ nm & $545.06\pm0.83$ nm & (H, C, O, N, Fe)\\
\hline
\citet{fasick98} & $488$ nm & $488.00\pm0.12$ nm & 11-cis retinal\\
Dolphin eye pigment & $545$ nm & $545.06\pm0.83$ nm & chromophore \\
\hline
  \caption{\textsl{Summary of the results.}
           The first column lists the found articles, with
           their topics.
           The second column shows the transition wavelengths reported
           in each article. 
           We quote the uncertainty when it is reported, otherwise we assume it is the last significant digit, that is 1 nm in most cases, which is the median uncertainty of published transition quoted in Table~\ref{tab:results}.
           The third column gives the closest DIB reported by \citet{hobbs09}.
           This database contains 380 bands in the $\lambda=380-810$~nm range.
           We do not discuss bands reported by the interdisciplinary articles outside this spectral range.
           Values in grey correspond to the case where the DIB centroid is not consistent with the article measurement.
           The last column gives the studied molecule, with its constituting elements between parentheses, when they are provided.}
  \label{tab:results}
\end{longtable}
The twelve interdisciplinary articles listed in Table~\ref{tab:results} are all biochemistry studies.
They all report experimental spectroscopic measurements of organic molecules.
Most of these molecules are constituted of abundant interstellar atoms, mainly, H, C, N and O.
We can divide them into the two following categories.

  \subsubsection{Chromophores}

\begin{figure}[htbp]
  \begin{tabular}{ccc}
    \includegraphics[width=0.29\textwidth]{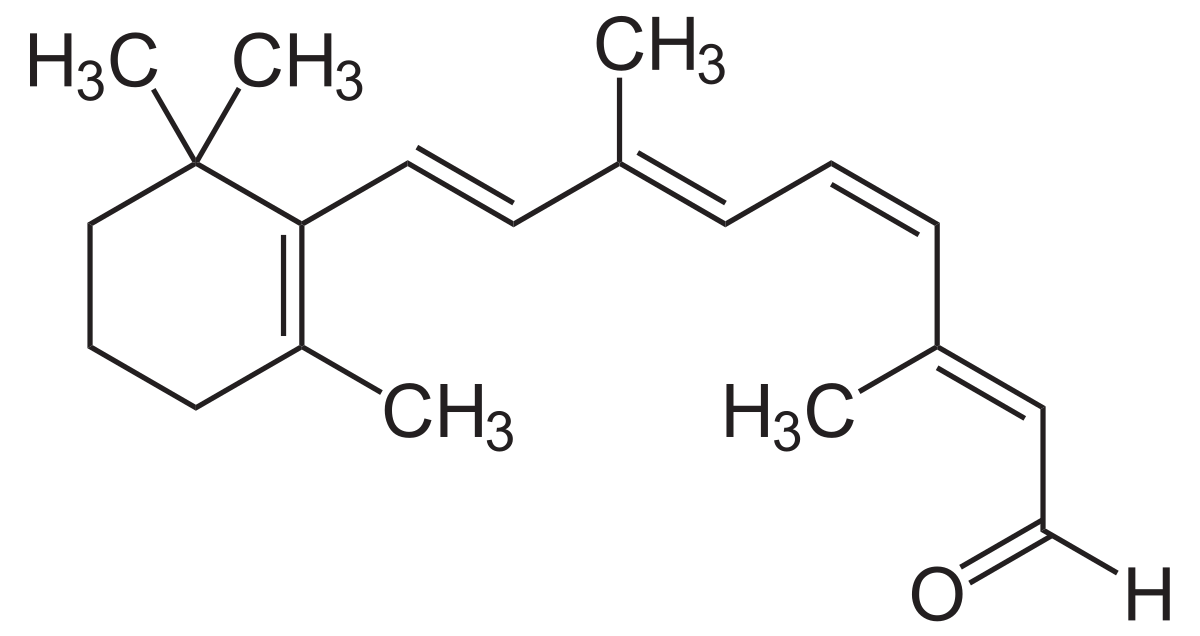} 
    &
    \includegraphics[width=0.26\textwidth]{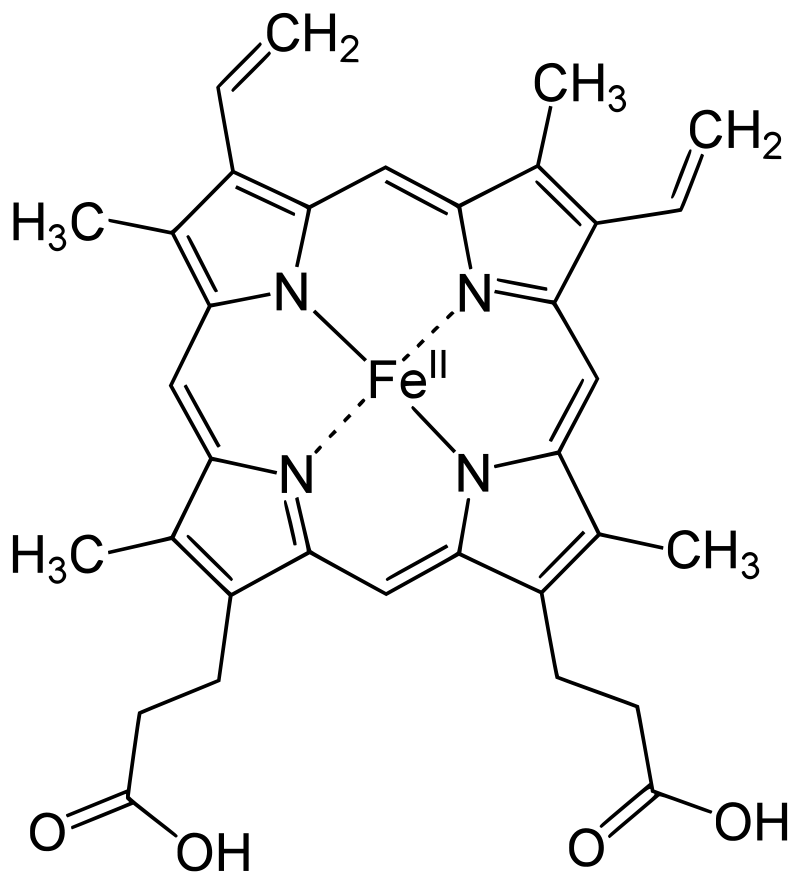} &
    \includegraphics[width=0.35\textwidth]{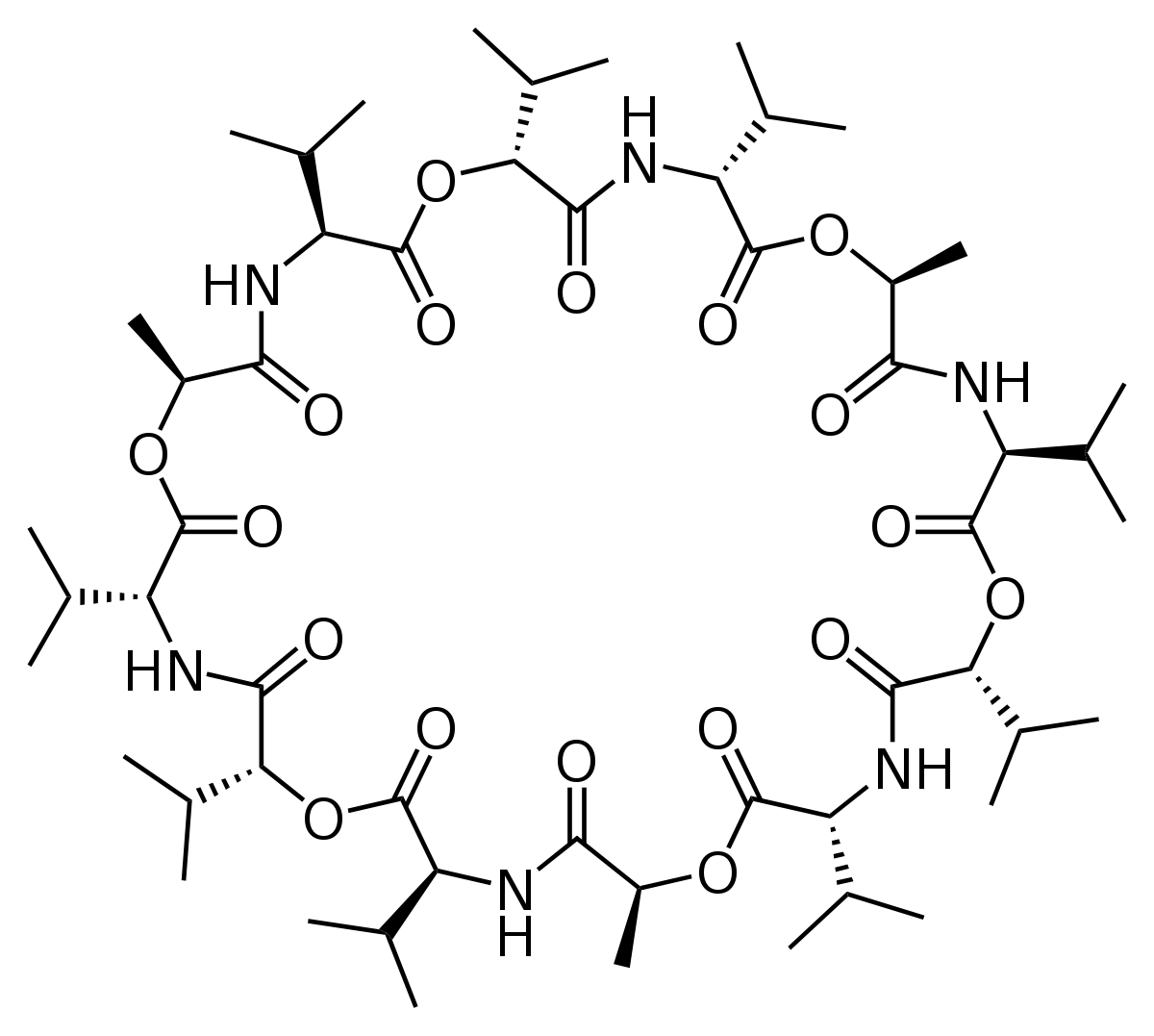}
    \\
    11-cis retinal & Heme b & Valinomycin \\
  \end{tabular}
  \caption{\textsl{Molecular structure of 11-cis retinal, heme b and valinomycin.}}
  \label{fig:molecules}
\end{figure}

A majority of the papers in Table~\ref{tab:results} deal with animal retinal eye pigments \citep{fasick98,spady06,davies07,davies09,wakakuwa10}.
These studies indeed measure the absorption bands of organic molecules in the visible range.
Their potential relevance to DIBs is thus obvious.
All these molecules are proteins (such as opsin) containing \textit{chromophores}.
Chromophores are molecular groupings, often part of a larger molecule, that are responsible for the color of an organism \citep[\eg][for a review]{shukla17}.
It is thus reasonable to assume that they are responsible for the reported bands.
A recurring chromophore in these studies is 11-cis retinal (Figure~\ref{fig:molecules}).
Not all these papers discuss the actual chromophores present in their molecules, probably because they are not always known.

  \subsubsection{Heme and other molecules}

Apart from chromophores, our model points us toward several other organic molecules.
In particular, two studies deal with molecules related to heme \citep[Figure~\ref{fig:molecules};][]{remigy03,marechal07}.
Heme is the molecule that allows hemoglobin to transport oxygen in the blood stream.
Another potentially interesting molecule for interstellar chemistry is valinomycin \citep[Figure~\ref{fig:molecules};][]{wolfbeis00}.
We will discuss the likeliness of the existence of these molecules in the ISM in Sect.~\ref{sec:discussion}.

\subsection{Discussion}

We now attempt at assessing these results.
We start by discussing the numerical values of the spectral band centroid.
We then evaluate the likeliness of these molecules as carriers of some DIBs.

  \subsubsection{Accuracy of the reported transitions}
  
The second column of Table~\ref{tab:results} lists the central wavelengths of the bands reported by the articles found by our model.
Unfortunately, only a few of these papers quote measurement errors on the centroid.
This adds some uncertainty to our assessment.
We assume that these errors are the last significant digit, that is 1~nm in most cases.
Yet, we note that the median of the uncertainties in Table~\ref{tab:results}, when they are reported, is exactly 1~nm.
Our assumption is thus realistic, assuming errors are comparable across all these studies.
The third column of Table~\ref{tab:results} gives the closest DIB from the measured band, using the \citet{hobbs09} compilation.
Out of the 43 centroids, only 8 ($19\,\%$) do not have a DIB within $\pm1\sigma$ (grey values).

The first question one can ask is how probable is it to draw a centroid wavelength at $\pm1\sigma$ of a DIB.
The compilation of \citet{hobbs09} contains 380 DIBs in the wavelength range $\lambda=320-810$~nm.
It gives a probability of 0.78 DIB per nm.
Assuming that DIBs are randomly, uniformly distributed with this density, the Poissonian probability to find at least one DIB within $\pm1\sigma$ ($\pm1$~nm in our case) is $79\,\%$.
Matching bands is thus very likely.
However, several studies list a large number of bands.
The most spectacular is the paper by \citet{davies09}, listing 11 bands with their uncertainties.
It happens that all of them coincide with a DIB within $\pm1\sigma$.
Using the uncertainties quoted by \citet{davies09} and assuming they are normally distributed and independent, the probability to randomly obtain such a result is $0.2\,\%$.
These results are thus non trivial. 

We have mentioned in Sect.~\ref{sec:DIBs} that DIBs were a few Angstrom wide.
The bands reported in these articles are however wider (up to several tens of nm).
One can thus wonder if they are relevant to our problem.
The answer to this question is not straightforward.
Molecular band width depends on the molecular size. 
Our candidate molecules contain a few tens to a hundred atoms, which is the expected size of DIB carriers \citep{macisaac22}.
If these molecules are however bonded in a larger matrix, such as a protein, the line widths will surely be broadened with respect to an isolated molecule. 
This will be because, unless the material is very pure and well structured, the local environment of the chromophore will be varied thus giving to slight spread in line positions due to different (steric or other) interactions for each molecule.

  \subsubsection{Insights for astrochemistry}
  \label{sec:discussion}

We have seen that most of the molecules our model points us toward are constituted of abundant interstellar atoms.
This is a first requirement that our model has successfully accounted for.
It is likely the result of the NLP training, associating DIBs with these elements and with organic molecules.

Our highest-score molecules, chromophores, have been proposed as DIB carriers \citep[\eg][]{johnson06,adams19}, and this is probably why our NLP model has selected them.
Chromophores are probably too brittle molecules to survive in the diffuse ISM.
For instance, the long chain of 11-cis retinal (Figure~\ref{fig:molecules}) will likely be photolyzed by UV photons.
However, chromophores could form moieties in larger molecules or, possibly, in nanometer-size dust grains \citep{jones14,jones16b}.
This would allow them to be present in the ISM and carry some of the DIBs.
We note this is also the case in the biochemistry studies. 
Several of the found papers mention that the chromophores are covalently bonded with their proteins.

Heme (Figure~\ref{fig:molecules}) has also been discussed has a possible interstellar molecule \citep{jones16a}.
Together with porphyrin, which is also considered as a potential DIB carrier \citep{johnson94}, these molecules could result from the reaction of nitrogen atoms with hydrogenated amorphous carbon grains.
The structure of valinomycin (Figure~\ref{fig:molecules}) is not unlike heme and porphyrins.
The problem of valinomycin is that it is probably not going to be very stable in the ISM because of all its oxygen, which is going to make it very reactive with atomic C, N, H and S.

\subsection{Limitation of the methodology and prospects for interdisciplinarity}

This pilot study aims to demonstrate a proof of concept on using NLP as a tool for users to identify cross-domain knowledge.
We have shown that it could bring down the borders between different disciplines and perform a rather unbiased search for relevant information.
Our uncovering of meaningful classes of molecules, studied in biochemistry, that could provide insights into an astrophysical question shows the practical relevance of NLP for interdisciplinary studies.
Could this method, applied to the particular problem of the origin of DIBs, be applied to other open questions? 
There are no reason why this could not be the case.
The method is however not completely straightforward and will need to be adapted to each new problem and its specificity: what is the size of the specialized corpus, are numerical quantities instrumental, and so on?
Fully automatizing the search in order to release an open source software is thus a challenge for future studies.

As the nature of our research question is closely associated to the DIBs wavelengths, these physical quantities are thus the ideal lampposts guiding us in the literature outside of the astrophysics domain.
To reproduce such process for research questions that are associated to some semantic concepts, an embeddings space of higher precision is crucial.
It is thus a natural next step to explore how one can disambiguate word senses in the embeddings space.
It is also worthwhile to explore other domain-adaptation techniques, such as the graph-based latent semantic imputation~\citep{yao_enhancing_2019}, to obtain an embeddings space with higher precision in the domain of interest.

As mentioned in Section~\ref{ref:w2v}, the study is based on a generic Word2Vec-CBOW model that is trained on 1.5 million English articles selected with some specific keywords.
These manually selected keywords inevitably create some domain biases in the generic corpus.
This can be further improved by randomizing our search across the entire CORE database, which contains more than 500 million articles.
Furthermore, we include only abstracts in our analysis in this study due to the limitation of resources.
A lot of important contextual information written in the full text is therefore absent from our analysis.
To thoroughly cover all information in the generic and DIB corpus, it would be essential to include the full text information in the future study.

Another limitation of the demonstrated method in this work is the language.
As the curated DIB corpus includes only English articles, for research questions that are in low-resource languages, it would be difficult to adopt such methodology in cross-domain literature search.

\section{SUMMARY AND CONCLUSION}
\label{sec:summary}

We have trained a \textit{Natural Language Processing} (NLP) model on a corpus of astrophysical publications dealing with the open question of the origin of \textit{Diffuse Interstellar Bands} (DIBs).
We have then used this model to explore an interdisciplinary corpus of scientific literature, with the hope to find relevant molecules having transitions matching some DIBs.
We have implemented a careful parsing of physical quantities and their units, in order to identify measured wavelengths in the visible electromagnetic spectrum.
Our model points us toward twelve biochemistry studies presenting spectroscopic measurements of molecules having transitions consistent with some DIBs. 
More than half of these molecules contain chromophores.
Several other studies deal with molecules linked to heme and valinomycin.

Our main objectives, the feasibility to use NLP to address open scientific questions, is reached.
We have shown that NLP is able to surface candidate DIBs molecules from an interdisciplinary corpus of scientific literature.
This confirms that NLP-methodologies can generate plausible and non-trivial hypotheses for future investigation.
First, we have shown that some associations between the reported transitions and the DIBs would be unlikely if they were purely random.
Second, our NLP model has pointed us toward molecules relevant to the \textit{InterStellar Medium} (ISM).
These molecules are constituted of abundant interstellar atoms.
In addition, several of these molecules have been proposed in the past to be DIB carriers.
Their presence in the ISM has not been proven, yet, but it is conjectured that they could form moieties in interstellar grains.
In light of our study, NLP thus appears as a practical tool for interdisciplinarity.
We have shown that this computational linguistic technique could uncover information in the biochemistry literature that happen to be relevant to astrophysics.
We have also discussed how this technique could be applied to other open questions in an interdisciplinary fashion.
In the future, extending our analysis to the full text, and expanding the generic corpus to thoroughly-cover cross-domain literature, can potentially give us a more complete result.
From an astrophysical point of view, our work adds credibility to the possibility that some DIBs could be carried by chromophores and to the role of heme and porphyrins in interstellar chemistry.
These findings need now to be further investigated in the laboratory, in interstellar conditions.
As a further outlook, our methodology should be applicable to other open scientific questions which require interdisciplinary knowledge.
More generally, our study shows it is possible to retrieve universal concepts from multiple disciplines. An interesting prospective would be to look for hidden interdisciplinary concepts from related universal concepts used in multiple domains.

\subsection*{Acknowledgements}
We thank Anthony Jones for a useful discussion about the molecules found by our model and about DIBs in general, and Jason  Hoelscher-Obermaier for constructive comments on the prospects of this study.
We also thank the two anonymous referees for their comments which improved the quality and clarity of this article.
This paper was supported by funding from NRC, Project ID: 309594, the AI Chemist under the collaboration of Iris.ai AS with Dr. Frédéric GALLIANO covering the contributions of Viktor BOTEV, Jeremy MINTON, Ronin WU, and partly Corentin VAN DEN BROEK D'OBRENAN.
This work was supported by the Programme National "Physique et Chimie du Milieu Interstellaire" (PCMI) of the CNRS/INSU with INC/INP co-funded by CEA and CNES.

\bibliographystyle{jimis-en}
\bibliography{references}

\end{document}